# An Intelligent-Detection Network for Handwritten Mathematical Expression Recognition

**Ziqi YE**
*School of Information Science and Technology, Fudan University, China*
*yezq21@m.fudan.edu.cn

**Abstract:** The use of artificial intelligence technology in education is growing rapidly, with increasing attention being paid to handwritten mathematical expression recognition (HMER) by researchers. However, many existing methods for HMER may fail to accurately read formulas with complex structures, as the attention results can be inaccurate due to illegible handwriting or large variations in writing styles. Our proposed Intelligent-Detection Network (IDN) for HMER differs from traditional encoder-decoder methods by utilizing object detection techniques. Specifically, we have developed an enhanced YOLOv7 network that can accurately detect both digital and symbolic objects. The detection results are then integrated into the bidirectional gated recurrent unit (BiGRU) and the baseline symbol relationship tree (BSRT) to determine the relationships between symbols and numbers. The experiments demonstrate that the proposed method outperforms those encoder-decoder networks in recognizing complex handwritten mathematical expressions. This is due to the precise detection of symbols and numbers. Our research has the potential to make valuable contributions to the field of HMER. This could be applied in various practical scenarios, such as assignment grading in schools and information entry of paper documents.

## 1. Introduction

Handwritten mathematical expression recognition (HMER) has multiple applications, including assignment grading, digital library services, human-computer interaction, and office automation. Offline HMER is generally considered more difficult than online HMER due to differences in handwriting styles and the complex two-dimensional structure of mathematical formulas (Zhang et al., 2018). Recently, the methods based on encoder-decoder structure have shown obvious success (Cho et al., 2014). However, these methods may not always ensure accurate attention, particularly when the structure of a handwritten formula is complicated or the mathematical notation is unclear (Li et al., 2022). To alleviate this problem, we propose the Intelligent-Detection Network (IDN) that uses improved object detection network YOLOv7 and a structural analysis module based on bidirectional gated recurrent unit (BiGRU) and baseline symbol relationship tree (BSRT) to obtain HMER results. Our extensive experiments demonstrate that IDN outperforms other excellent methods on the HME100K dataset.

## 2. Related Work

*2.1 HMER*

HMER methods have been comprised of three key components: symbol segmentation, symbol recognition and structure analysis (Chan & Yeung, 2000). Convolutional neural network (CNN) are capable of learning and automatically extracting features from original image data (Krizhevsky et al., 2017). These low-level feature representations are transformed into high-level feature representations, allowing CNN to effectively complete complex tasks within the field of computer vision. To alleviate the problem of inadequate coverage, Zhang et al. (2017) proposed the WAP encoder-decoder model, which consists of a CNN encoder based on visual geometry group (VGG) architecture and a decoder that uses a recurrent neural network (RNN) with gated recurrent unit (GRU). The WAP model encodes the input image of a mathematical formula to extract high-dimensional features and then decodes the corresponding LaTeX sequence of the formula.

In the subsequent research, Zhang et al. (2018) proposed the DenseWAP model to enhance the performance of CNN. The model replaces the VGG in the WAP model with the dense convolutional neural network (DenseNet). Wu et al. (2019) proposed the PAL model and PAL-V2 model which combine deep learning and adversarial learning to overcome the change of writing style. The PAL-V2 model uses a decoder based on CNN to solve the problems of gradient disappearance and gradient explosion in RNN (Wu et al., 2020).

## 2.2 Object Detection

Object detection algorithms can be broadly classified into two categories: two-stage algorithms and one-stage algorithms. Two-stage algorithms identify regions of interest and then analyze the contents of these regions to obtain detection results, while one-stage algorithms directly perform regression analysis to obtain detection results without explicitly identifying regions of interest. Ren et al. (2015) proposed the Faster R-CNN algorithm, which replaces selective search with a regional suggestion network, thereby significantly improving training efficiency. However, while the detection effect of two-stage object detection algorithms is improving, the detection speed remains limited by the backbone network structure.

Redmond et al. (2016) proposed the one-stage object detection algorithm called you only look once (YOLO) for improving detection speed in practical application scenarios. YOLO has a simple training process and fast detection speed compared to two-stage object detection algorithms. However, it has limitations in detecting small objects and can only detect one object at a time. YOLOv3 model introduced the feature pyramid technique to obtain multi-scale features and balance detection speed and accuracy by changing the model's structure (Redmon & Farhadi, 2018). YOLOv4 improved the model's detection capabilities by incorporating the CSPDarknet-53 backbone network and Mish activation function (Bochkovskiy et al., 2020). The recognition ability of the network and the generalization of the network have been improved in YOLOv7 by incorporating the efficient layer aggregation network (ELAN) module and SPPCSPC module (Wang et al., 2022). As a result, YOLOv7 is currently the most advanced algorithm for one-stage object detection.

## 3. Methods

### 3.1 Intelligent-Detection Network Based on improved YOLOv7

As shown in Figure 1, we present our Intelligent Detection Network (IDN) which is an end-to-end trainable architecture consisting of the improved YOLOv7 network, BiGRU and BSRT. To further enhance the feature extraction of numbers and symbols, we incorporate a feature enhancement module (FEM) into the YOLOv7 structure. FEM uses extended convolution to learn feature information at different scales in order to improve the accuracy of multi-scale target detection and recognition (Wang et al., 2022). We use the feature fusion technique of FEM to enhance the accuracy of multi-scale predictions in HMER.

The input image is first processed by the backbone which consists of the ELAN and MP-1 modules. The resulting feature is then passed through a combination of modules including SPPCSPC, MP-2, FEM and others in the head stage to transmit and locate the feature information. Finally, the module based on BiGRU and BSRT is used for structural analysis to obtain the recognition result.

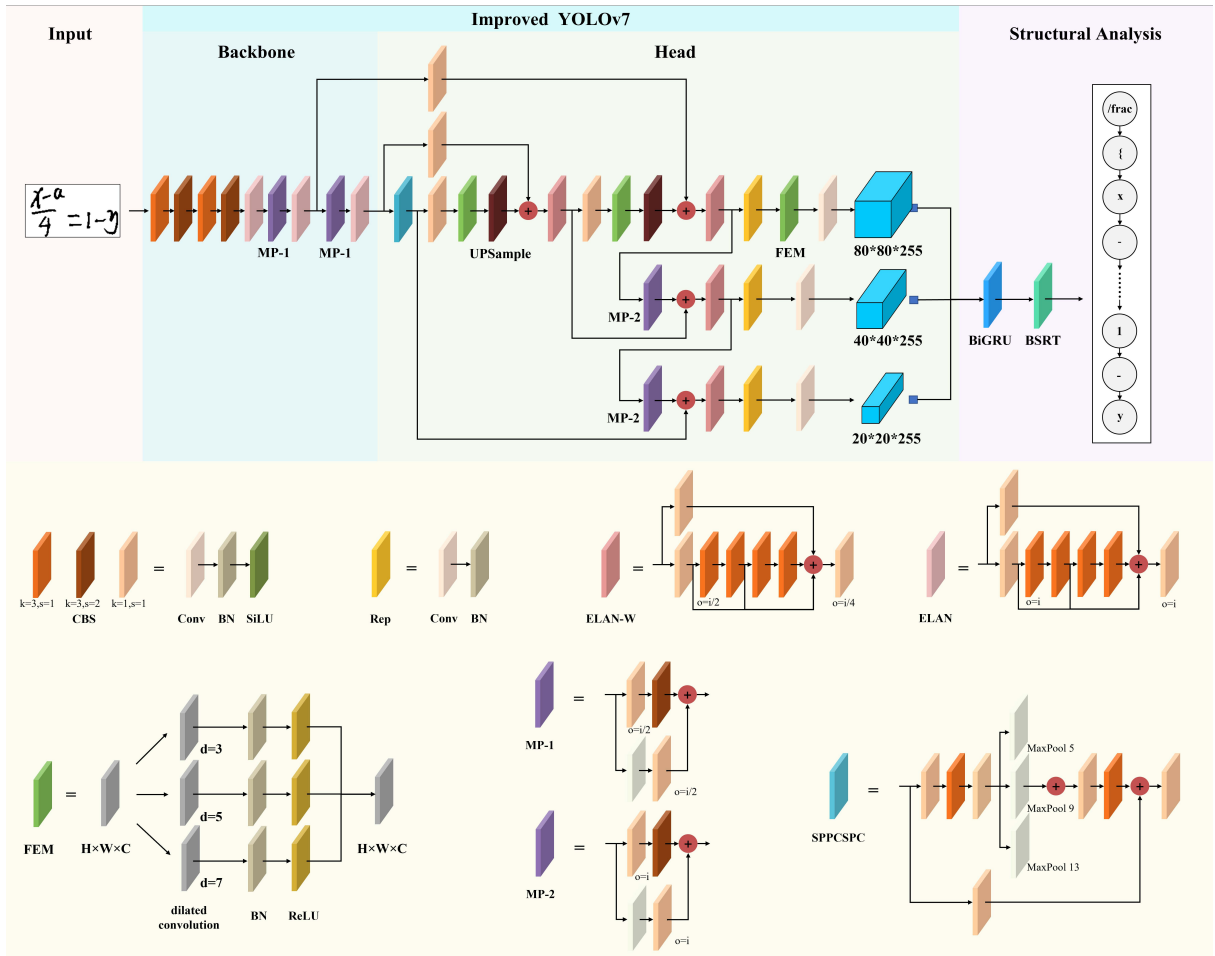

*Figure 1.* Structure of the Proposed Intelligent-Detection Network (IDN).

### 3.2 Structural Analysis Module Based on BiGRU and BSRT

To improve information processing efficiency and reduce model parameters and tensors, GRUs replace the three gated cycle units of long short-term memory (LSTM) with reset gate ($r_t$) and update gate ($z_t$). This makes GRUs a more concise and efficient option compared to LSTM. The BiGRU layer takes the vector matrix from the previous layer as input and uses a bidirectional gated recurrent neural network to extract semantic information and feature structure. Unlike a regular GRU which only considers past information, the BiGRU consists of forward GRUs and backward GRUs that can consider both past and future information.

Figure 2 illustrates the structure of GRU and BiGRU. We use BiGRU to extract symbol category and position information from the improved YOLOv7 detection results. This extracted information is then fed into the BSRT for final processing.

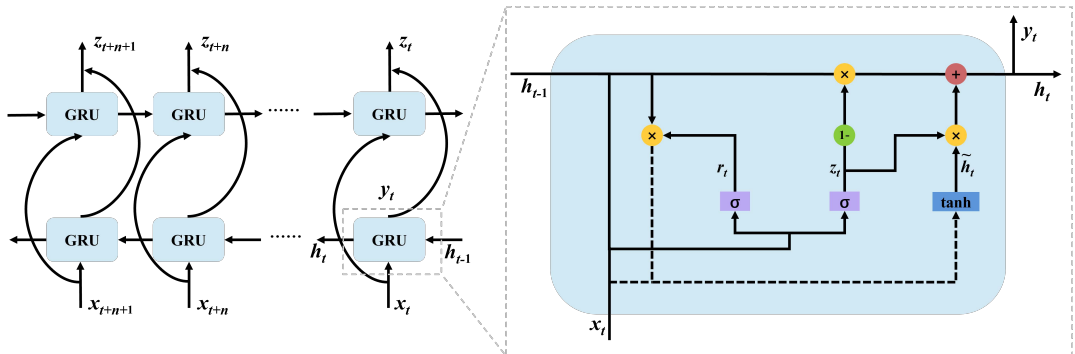

*Figure 2.* Structure of the BiGRU and GRU in IDN.

Last but not least, we use BSRT algorithm to produce mathematical expression results in structural analysis module. BSRT defines seven spatial relationships, including above, below, right, superscript, subscript, and unrelated. After using the improved YOLOv7 algorithm for object detection and recognition, the center point, length, and width of symbols within formula species are identified. This allows for the determination of seven distinct spatial relations between each symbol in the formula based on the gathered information.

However, obtaining precise formula structures cannot be achieved solely through the position information of each symbol. To address this, we also consider the center offset, aspect ratio, and overlap interval ratio of the symbol. The center offset θ is determined by measuring the angle between the line at the center point of the adjacent symbol and the horizontal line. The aspect ratio α and β of a symbol are calculated by dividing the length of the symbol by the length of the adjacent symbol and the width of the symbol by the width of the adjacent symbol, respectively. The overlap interval ratio λ and μ of a symbol are expressed as the ratio of the projected length of the two symbols in the vertical or horizontal direction to the length or width of the previous symbol. The relationship between symbols is determined using the three functions that have been defined. For example, if the center offset θ between two symbols is between 0.4π and 0.6π and the overlap interval ratio μ is greater than 0.5, the upper and lower relation between the two symbols is considered to be satisfied.

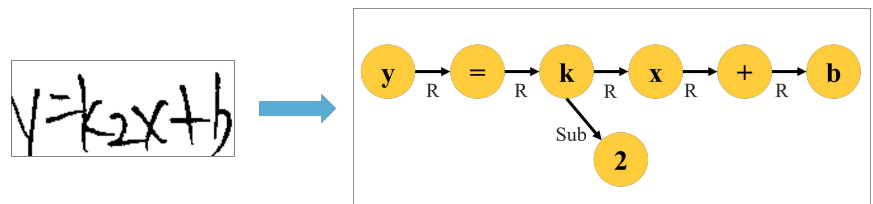

*Figure 3.* A BSRT Generated From an Image.

Figure 3 gives the resulting BSRT structure following object detection and recognition. In BSRT structure, each node represents a symbol, and the label on each side denotes the relationship between the two symbols.

## 4. Experiments and Results

### 4.1 Datasets

We select the offline dataset HME 100K as our experimental dataset (Yuan et al., 2022). The HME100K dataset is a collection of handwritten mathematical expressions from real-life scenarios, comprising 74,502 training images and 24,607 testing images. These images are sourced from thousands of photos of students' paper, and they contain real background and color information. Therefore, we need to binarize them before inputting them into the network to facilitate subsequent detection and recognition.

### 4.2 Evaluation Metrics

We evaluate the performance of HMER using the expression recognition rate (ExpRate) as the evaluation metrics. ExpRate is calculated by determining the ratio of correctly recognized symbols to all symbols. To better assess the effectiveness of the algorithm, we also consider the tolerance levels of ≤1 and ≤2, which indicate that the expression recognition rate can tolerate one or two sign-level errors at most.

### 4.3 Results

The proposed IDN is implemented in PyTorch 1.10 and trained on a single Nvidia RTX 3090 with 24GB RAM. Figure 4 gives some results identified by IDN algorithm, including the identified LaTeX sequence results and visualization results.

| Input Image | Recognition Result | Visualization |
|---|---|---|
| 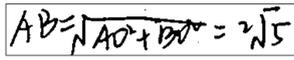 | AB=\sqrt{AO^2+BO}=\sqrt[2]{5} | $AB = \sqrt{AO^2 + BO^2} = \sqrt[2]{5}$ |
| 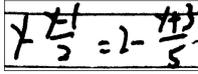 | y-\frac{y-1}{2}=2-\frac{y+3}{5} | $y - \frac{y-1}{2} = 2 - \frac{y+3}{5}$ |
| 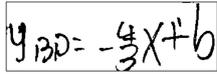 | y_{BD}=- \frac{4}{3}X+b | $y_{BD} = -\frac{4}{3}X + b$ |
| 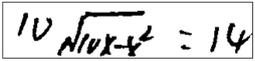 | 10\sqrt{10x-x^2}=14 | $10\sqrt{10x - x^2} = 14$ |
| 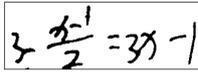 | 3- \frac{x-1}{2}=3x-1 | $3 - \frac{x-1}{2} = 3x - 1$ |

*Figure 4.* Some Recognition Results of IDN.

The results of our experiments indicate that the proposed IDN algorithm has achieved a recognition accuracy ExpRate of 67.82%. It increases to 82.91% when allowing for 1 symbol error and further increases to 88.37% when allowing for 2 symbols errors.

To demonstrate the effectiveness of our algorithm, we conducted experiments comparing it to other excellent algorithms in the same environment. As shown in Table 1, our IDN algorithm achieved an increase in ExpRate of 1.89% compared to ABM, 5.97% compared to DWAP, and 10.6% compared to PAL-v2. Furthermore, IDN outperformed the state-of-the-art algorithm ABM for ExpRate, ≤1 and ≤2.

Table 1. *Experimental Results of IDN and Other Algorithms*

| Method | ExpRate | ≤1 | ≤2 |
|---|---|---|---|
| WAP | 56.58% | 68.14% | 74.26% |
| PAL-v2 | 57.22% | 69.21% | 75.37% |
| DWAP | 61.85% | 70.63% | 77.14% |
| ABM | 65.93% | 81.16% | 87.86% |
| IDN (ours) | 67.82% | 82.91% | 88.37% |

To better understand the benefits of our proposed algorithm IDN, we provide some examples of its performance compared to DWAP in the HMER task. Figure 5 demonstrates that IDN is capable of recognizing intricate symbols, such as the "x" the first image, the "-" in the second and third images, and the number "3" in the fourth image.

| Input Image | DWAP | IDN (Ours) |
|---|---|---|
| 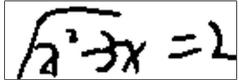 | \sqrt{a^2-3x}=2 | \sqrt{x^2-3x}=2 |
| 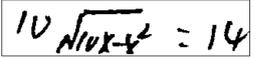 | 10\sqrt{10x+x^2}=14 | 10\sqrt{10x-x^2}=14 |
| 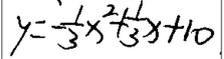 | y=\frac{1}{3} x^2+ \frac{1}{3} x+10 | y=-\frac{1}{3} x^2+ \frac{1}{3} x+10 |
| 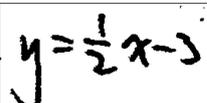 | y=\frac{1}{2} x-J | y=\frac{1}{2} x-3 |

*Figure 5.* The Comparison of Recognition Results Between IDN and DWAP.

# 5. Conclusion

This paper presents the design of an intelligent detection network named IDN that uses object detection and structure analysis to achieve better recognition performance compared to other excellent HMER algorithms. The proposed IDN incorporates a FEM module to adaptively learn feature information from multiple receptive domains, thereby enhancing the detection performance of the original YOLOv7 algorithm. In addition, we developed a structure analysis module, which uses BiGRU and BSRT, to determine the recognition results of the final formula. Our experiments demonstrate that the proposed IDN algorithm effectively recognizes handwritten mathematical expressions and outperforms other existing algorithms.